\documentclass[11pt]{article}

\usepackage[preprint]{acl}

\usepackage{times}
\usepackage{latexsym}

\usepackage[T1]{fontenc}

\usepackage[utf8]{inputenc}

\usepackage{microtype}

\usepackage{inconsolata}

\usepackage{graphicx}

\title{Logic of Montage}

\author{
Hayami Takahashi$^{1}$\thanks{\,\,Corresponding author.},
Kensuke Takahashi$^{2}$\\
$^1$Cach\'{e} et Wavelet, Inc. , Yokohama, Japan\\
$^2$Department of Mathematics, Tokyo University of Science , Shinjuku-ku, Tokyo, Japan\\
{\small \texttt{hayami.takahashi@cachewavelet.jp} }, \ {\small \texttt{1125092@ed.tus.ac.jp}}
}

\begin{document}
\maketitle

\begin{abstract}

In expressing emotions, as an expression form separate from natural language,
we propose an alternative form that complements natural language, acting as a proxy or window for emotional states.
First, we set up an expression form "Effect of Contradictory Structure(EoCoS)."
EoCoS is not static but dynamic.
Effect in EoCoS is unpleasant or pleasant, and the orientation to avoid that unpleasantness is considered pseudo-expression of will.
Second, EoCoSs can be overlapped with each other.
This overlapping operation is called "montage."
A broader "Structure" that includes related EoCoSs and  "Effect of Structure (EoS)" are set up.
Montage produces EoS.
This process is called "EoS process."
In montage, it is necessary to set something like "strength," so we adopted Deleuze and Deleuze/Guattari(DG)'s word "intensity" and set it as an element of our model.
We set up a general theoretical framework - Word Import(Transplant,Graft) Between Systems (Models) -  and justified the import of "intensity" through Austin's use of the word "force."
EoS process is demonstrated using the example of proceeding to the next level of education. \footnote{\url{https://github.com/iwatesans/montage}}

\end{abstract}

\section{Introduction}

\citet{Natsume:1908} wrote, 
"How can we explain the emotions we derive from a literary work using tangible words or symbols?" 
 "The only way is to break it down."  
"Breaking down emotions is extremely difficult." 

This difficulty remains unchanged to this day.
We propose a conceptual tool to alleviate this difficulty and opacity.
This tool consists of three elements: "Effect of Contradictory Structure(EoCoS)," montage, and "Effect of Structure(EoS)."

EoS process is expressed in two stages.
First, "Effect of Contradictory Structure (EoCoS)" is expressed (Figure 1).
EoCoS is considered to be able to express contradictions.
When in a contradictory state, EoCoS is in a state of unpleasant and aims to cross border to avoid unpleasant, 
while when not in a contradictory state, it aims not to cross border.
To emphasize this aspect, it can also be called "Cross Border Form (CBF)."
"Orientation to avoid unpleasant" can be considered a pseudo-expression of "will."
"Intensity" is set up in EoCoS.

\begin{figure}[h]
\begin{center}
  \includegraphics[width=60mm]{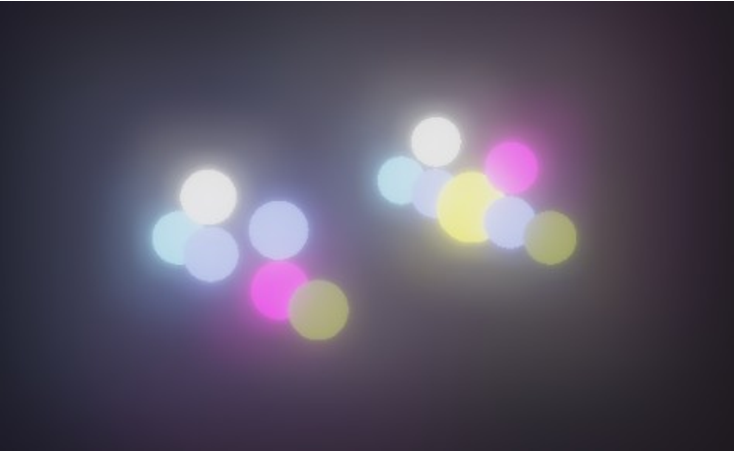}
  \caption {Effect of Contradictory Structure (EoCoS)}
\end{center}
\end{figure}

Second, EoCoSs are superimposed on each other.This superimposition process is called "montage."
In montage, the phenomenon of "sameness" occupies a privileged position.
The following is set up:
An "action" occurs between "same" EoCoS, and they are "strengthened."
And between "opposite" EoCoS within the same subject, "action" also occurs.
In EoS after montage, by setting up influence thresholds, the scope of effect can be made clear.

"Intensity" is imported from the model (system) of \citet{Deleuze:1968} and \citet{Deleuze:1980} and is closely related to the word "force."
\citet{Deleuze:1986} organized \citet{Foucault:1975}'s theory about "micro-power."
Incorporating Austin, Foucault's approach to prison, \citet{Hjelmslev:1943}'s "form of content" and Stoic'theory, \citet{Deleuze:1980} constructed a new linguistics.
The model of Deleuze and DG contain many neologisms.
We need to verify whether "intensity" can be imported (transplanted, grafted) into our model without conflicting with these neologisms.
Such verification forms a general theoretical framework - Word Import Between Systems.
We investigate the correspondence between Austin's "illocutionary force" and Deleuze's "map of relations between forces."

In summary, our contributions are as follows:
\vskip\baselineskip
\begin{itemize}
\item We propose an expression form EoCoS which makes emotions more visible.
\item EoCoS can be considered as pseudo-expression of will.
\item Montage of EoCoSs is able to clarify the scope of effect (e.g., the scope of influence of demonstrative pronouns).
\item We introduce a general theoretical framework - Word Import(Transplant,Graft) Between Sytems(Models). 
\item We investigate the similarity between Austin's "force" and Deleuze's "force."
\end{itemize}

\vskip\baselineskip
\section{Ability to express emotional states}
We examine three examples and one metaphor to see whether or not it is necessary to express emotional states.

\vskip\baselineskip
-- Example 1 --

Below is the Spanish folktale "Fox and Chicken" \citep{NHK:2022}.

\begin{quote}

One day, a chicken saw a fox approaching him. To escape, the chicken climbed a tree and croaked.
"Cock-a-doodle-doo!"
The fox looked up toward the sound and asked the chicken, "Why did you climb so high? Are you scared of me?"
"Yes, yes."
"Well, there's no need to be scared of me. Because, a new rule has just been put in place to help us all get along without hurting each other. 
Don't you know $\underline{\emph{\textbf{that}}}_{\,01}$ ?"
"No, I don't."
"Then, come down. I'll tell you $\underline{\emph{\textbf{that}}}_{\,02}$ ."

\end{quote}

Set the following:

\vskip\baselineskip
- A -

(1) make the fox speak more politely.

(2) make the fox a young child, or, taking the anthropomorphism further, make the fox a legal scholar.

(3) make the fox's speech more rough.

- B -

This story is read by two people, P and Q, and P asks Q the following questions.

(1) What does $\underline{\emph{\textbf{that}}}_{\,01}$  refer to? (What is the content of $\underline{\emph{\textbf{that}}}_{\,01}$ ?)

(2) What does $\underline{\emph{\textbf{that}}}_{\,02}$  refer to? (What is the content of $\underline{\emph{\textbf{that}}}_{\,02}$  ?)

- C -

What will be "Q's mental state" when P asks Q the question B?
Note that "Q's mental state" refers to everything that occurred as a result of P's question.

\vskip\baselineskip
Around the verb "know" in the fox's utterance, "Don't you know about $\underline{\emph{\textbf{that}}}_{\,01}$  ?"
, the more rough the fox's speech becomes,
the more "unpleasant" becomes prominent due to some kind of "contradiction."
In our daily lives, not "knowing" about an enacted law is generally an "unpleasant" state.
The person may also be blamed.
The $\underline{\emph{\textbf{that}}}_{\,01}$  is the object of "know," very close to the state of "know."
Therefore, $\underline{\emph{\textbf{that}}}_{\,01}$  is bound within the realm of the "contradiction" involving "know."
And could it be said that $\underline{\emph{\textbf{that}}}_{\,01}$  is bound by the "intensity" of the "contradiction" involving "know"? When a fox speaks very politely, or is a legal scholar, the intensity of the contradiction involving "know" may be reduced.

Natural language does not have a hook that can directly and explicitly describe the differences in effects of changes in (1), (2), and (3) of setting A.
This characteristic is thought to be an indicator of the ability to express emotions.

\vskip\baselineskip
-- Example 2 --

Picture the following story in the form of a conversation.

\begin{quote}

A family goes on a trip together once a year.
The parents promise to take their toddler to an amusement park.
Just before the trip, the parents received the passing of a close relative, and the trip had to be cancelled for the funeral.
To the toddler, whatever the reason for cancellation, that's just a promise broken.
When the parents told their toddler that they couldn't go on the trip, the toddler burst into tears and said, "Breaking a promise is bad, and you're a liar. I hate you."
The parents had a hard time calming their toddler.

\end{quote}

There are seven utterances:

u1. promise to take their toddler to an amusement park

u2. receive news of the death of a close relative

u3. tell we can't go to an amusement park

u4. you're wrong for breaking our promise

u5. you're a liar

u6. I hate you

u7. calm their toddler

\vskip\baselineskip
For the seven utterances, we can distinguish between the speaker and the listener, and between the feelings before and after the utterance, there are 7 x 2 x 2 = 28 feelings.

This story contains a promise.
The trip cancellation is not unreasonable because it involves the death of a person.
This story involves differing priorities.
In this difference,  emotions play a central role, and the social system is deeply involved.

Before u3, when we think "the trip cancellation is unavoidable,"
the notion "unavoidable" may be perceived as a real existence, but it may be a fabrication created by grammatical influence.
If we think of "unavoidable" as a situation, then that "unavoidable" may represent, or be an effect of, the tension between the toddler's sadness and the funeral.

In this event, what are the different elements involved? 
Can the above "unavoidable" be counted as an element?
Are those elements finite?
Is it possible to list all the elements involved and the relation between those elements?

\vskip\baselineskip
-- Example 3 --

The following is taken from \citet{Derrida:1980}'s essay discussing Poe "The Purloined Letter."

\begin{quote}

It was a freak of fancy in my friend (for what else shall I call it?) to 
be enamored of the Night for her own sake; and into this bizarrerie,  as 
into all his others, I quietly fell; giving myself up to his wild whims 
with a perfect abandon. The sable divinity would not herself dwell 
with us always; but we could counterfeit her presence. 

\end{quote}

How do readers perceive the repetition of words that describe "strangeness" ?
Does this repetition of "strangeness" contribute anything ?

It is not simply a repetition of "my" friend's "strangeness," but it is accompanied by the image of "me" being drawn into that strangeness.
To be drawn in means that "I" have crossed a border into the direction of what is pleasant.
This reiterate, which begins with "love for the night," is on the pleasant side.
Does the intensity of the night grow stronger ? 
Does the intensity grow until they finally counterfeit the night to keep it close ?
Does the repetition of "strangeness" prepare readers fully for perceiving the counterfeit ?

\vskip\baselineskip
-- Metaphor 1 --

Picture five ladles.
The ladles contain water, but in different amounts.
When we make a promise, the water in one ladle represents the effort required to fulfill the promise, and the water in the second ladle represents the direct disadvantages of failing to fulfill it.
Other factors are also involved in a promise.
Criticism when promises are not fulfilled, the direct benefits received when a promise is fulfilled, and the gratitude and praise given by those involved for fulfilling the promise.
These five ladles can sometimes exist simultaneously.
Criticism for failure cannot be separated from the relationship between the critic and the person being criticized, including their feelings towards each other.
Therefore, simply expressing negative emotions alone is insufficient to fully convey the criticism.

\vskip\baselineskip

Why did natural language settle into ("result in") this form? 
Given the constraints of this world, is there an abstract gradient involving labor-saving, visual visibility, and ease of settling? 
If an unsettling form exists, would its expressive power and expressive efficiency be the same? 
Does such expression form exist?

\section{Effect of Contradictory Structure}

\subsection{Cross Border Form}

EoCoS has the structure shown in Figure 1, 
and includes the following items: 
subject, object, aspects of the subject, aspects of the object, and a "pleasant" term.

EoCoS is a module that performs a reference(comparison, matching) between an "ideal image" and an "actual image."
The process from reference(comparison, matching) to resulting "effect" is considered a single set within this module.
As shown in the video \footnote{\url{https://github.com/iwatesans/montage}}, this module is not static but dynamic.

EoCoS is strongly based on the motivation to directly express "perception of difference." 
This is based on Saussure's hypothesis that the origin of meaning is "perception of difference."
From the standpoint of the rights, when "perceiving differences," 
it is fundamentally necessary to first compare the two entities.

EoCoS is divided into two parts by a border, and all items are positioned on one side or the other of the border ("biased positioning").
All items are able to transition across border.
When we emphasize the aspect of crossing border, we can also call EoCoS "Cross Border Form."

We define "contiguity" as when items are on the same side of the border.
If the object is something that is desired,  [pleasant] marker is placed in "contiguity" to it.

When the items are in the desired "contiguity" to each other,
the diagram is called "ideal image."
And the diagram showing the current state is called "actual image." 
The operation of comparing the two images is set up.
If they match, the result is "pleasant," and if they don't match, the result is "unpleasant."

\subsection{Expression of contradiction}

"Effect" primarily consists of either "pleasant" or "unpleasant" feeling, which arises from comparison(reference) between ideal image and actual image.
However, when a certain EoCoS is in a relationship of contiguity,  resemblance, or causation with another EoCoS, an interaction exists simultaneously.
Therefore, effect becomes a kind of mixture.

In EoCoS, when comparing ideal image and actual image, 
the inability to identify them, that is, the lack of identification, is considered to express a contradiction.

\subsection{Minimum configuration}

To minimize the module - EoCoS -, we considered how many types of relation exist. 
We tentatively adopted \citet{Hume:1739}'s hypothesis: three: resemblance, contiguity, and causation.
By constructing the module with elements that are as independent of each other as possible, similar to "basis" in signal processing,
we may be able to simplify and minimize the module.

\subsection{Pseudo-expression of will}
We consider the orientation to avoid unpleasant, which is a effect of contradictory structure, as a pseudo-expression of will (Appendix J).

If the will can be segmented (separated or extracted) from events, it can be inferred as part of a sequence of events. 
Such a form would allow for inference similar to that used in natural language processing.

\subsection{Individual differences}

EoCoS also has an affinity for modeling private language use, miscommunication, and argument.

The shape of the EoCoS unit(component) may be justifiable from the perspective of individual differences.
How can we model the unusual inclinations of a particular individual, inclinations that are not typical and exist only in a very small number of people ? 
Do the desires share the common characteristic of being a state of contradiction in which desires are not satisfied ?
Do they also share the commonality of harboring contradictions and seeking to transcend(cross) boundary(border) ?
Is the only difference the specific content to cross border ?

A person in a critical condition due to illness, struggling to breathe, might, in their distress, strangely connect one event to another and experience profound sadness.
The associations within that person's mind may seem bizarre, but the fact that such powerful mental images arose during their suffering cannot be denied.
For that person, that sadness is very real.
The dreams we have while sleeping at night may have some similarities to this example.

\subsubsection*{Conventional act}
Individual differences and a phenomena that are not conventional act partially overlap.

There are differing opinions about the range of conventional act.
\citet{Strawson:1964} and "Prolegomena(1967)"\citep{Grice:1989} are opposed to \citet{Austin:1962}'s assertion that "illocutionary acts are conventional acts" (Appendix G).
The word "conventional" also is related to "Use Theory of Meaning" which is compatible with current LLMs.

In EoCoS, we consider how many people hold a certain shape created by "biased positioning" of "pleasant" with other items.

\subsection{Intensity 1}

The term "intensity" is imported from Deleuze and DG, and we set it as an element of our model.
In the works of Deleuze and DG,
the concept of "intensity" has undergone changes\citep{Takahashi:2025b} in its usage in 1968, 1972, 1976-1980.
\citet{Deleuze:1968} disagreed with Bergson's "intensity"(Appendix K).
We deal with the "intensity" of 1968, 1976-1980.

The intensity of EoCoS is set to vary depending on the event.
When being free from contradiction, the state can be considered as one state of "intensity."
For demo, a provisional intensity is set.
The intensity should be adjusted to the appropriate level based on subsequent investigations.

\section{Montage}

The operation of superimposing EoCoS on one another is called "montage." 
This operation produces certain "effects." 
For the time being, we focus on the "strengthening" and "crossing border" that occur through superimposition.

\subsection{Eisenstein's montage}

Eisenstein's montage  is based on two elements:

\begin{quote}
I  believe  that  it  is  in  the  existence  of  these  two  elements  
- the  specific instance  of depiction  and the generalising image which pervades it - 
that the implacability and the all-devouring force  of artistic composition resides.

"Montage 1937"\citep{Eisenstein:1937}
\end{quote}

Eisenstein also emphasizes the simultaneity of the fragments.

\begin{quote}
I  have  written  and  spoken  many  times  about  montage  as being  not  so
much the sequence of segments as their simultaneity.
\end{quote}

Eisenstein makes several references to the use of montage in scenes from novels. 
For Eisenstein, montage is not something confined solely to film.

Within the framework of film theory, the concept of "montage" is a production technique (method).
There is a conflict between "montage" and "realism"(Appendix I) in terms of the granularity of technical classification. 
However, arguing that even within a single shot on the "realism" side, many subtle shifts in meaning occur.
we adapt "montage" as something more subtle.

\subsection{Intensity 2}

We set the following:
Two similar EEoCSs, when superimposed, result in "strengthening."
"Strengthening" means that the "intensity," which is the "degree" of "strength," increases.

Two EoCoSs within a causal chain, interact with each other, resulting in a change in intensity.

When considering the scope of effect, thresholds are set up to delimit the range of influence for changes in intensity.

\subsection{Action receptivity}

We set up that "strengthen" and "cross border" occurs only through "action."

When it is possible to say that one module can exert an action ("actionability") 
and the other module can be acted upon ("action receptivity"),
a "property that can be modified in intensity" is placed "on and between" the two modules. 
It is a property that appears simultaneously in both modules.

For the time being, 
we don't place a setting similar to "force propagation," 
which evokes the image of something propagating from one module to the other.

\subsection{Action point}
Regarding relation, Hume distinguishes between cases and emphasizes three: resemblance, contiguity, and causation.
We set it up as follows: When two EoCoSs are in any of these three relations, the connection point becomes the action point.

In demo, two Similar EoCoSs, or two EoCoSs within a causal chain, interact with each other, resulting in a change in intensity.

If we take DG's use of the word "intensity," 
we could say that "when there is a relation (resemblance, contiguity, causation) between two terms, intensity passes between the two terms."
But DG does not seem to say this. 
Furthermore, as mentioned above, our model will not set up something like "force propagation" for the time being.

\subsection{EoS process}
EoS processes is demonstrated using the example of proceeding to the next level of education (Appendix A).

\subsubsection*{Process characteristics}
Looking at demo , some characteristics can be observed.

(1) A border is set, and "pleasant" is positioned on one side or the other of that border. 
This expresses the biased positioning of "pleasant."
This bias expresses a contradiction.

(2) Within EoCoS, situations that are in a state of contradiction become the starting point for change.
In other words, the expression of change is related to the ability to express contradictions.

(3) Two Similar EoCoSs, or two EoCoSs within a causal chain, interact with each other, resulting in a change in intensity.
The concepts of "resemblance" and "causation" , which Hume emphasized as important in understanding relationships, serve as the starting points for changes in intensity.

\subsubsection*{Visualization of montage}
We are preparing to run EoS process on the machine.
There are seven steps to creating demo:
\begin{enumerate}
\item In order to make it easier to transform natural language into EoCoS, we convert natural language into another form, natural language 2 (NL2).

\item We transform NL2 into EoCoS. At that time, we make EoCoS into a format that is easy to montage.

\item For causation(causality), which is one of the relations, we classify causation and determine intensity which occurs between EoCoSs within causation.

\item For resemblance(similarity), which is one of the relations, we determine intensity.

\item Machine deploys montage of EoCoSs.

\item Machine calculates intensity for each montage.

\item Machine draws two: plates of EoCoSs , lines between EoCoSs.
\end{enumerate}

\vskip\baselineskip
We plan to create 1. and 2. through the LLMs.
Intensity in 3. and 4. needs to be updated to an appropriate value based on experimental science findings.
The shape of EoCoS is constantly being revised, and the demo shows the current shape.

EoS process is expected to be carried out by LLMs.

\section{Need for intensity monitoring}

\subsection*{Hesitation}

Hesitation, conflict, and distress all seem to require monitoring of "intensity." 
Hesitation arises when a subject is forced to choose between two options. 
It is a situation in which a smooth choice is not possible. 

Hesitation can sometimes arise after going through a process of strengthening emotions.
In such cases, intensity determines whether the logic branches into a state of hesitation or a smooth state without hesitation.

\subsection*{Values [Ricoeur]}

\citet{Ricoeur:1950} states that value reveals itself only to the extent that one is committed to it.

\begin{quote}
 Values  only  appear  to  me  in  proportion  to  my  loyalty, 
that  is,  my  active  dedication. 
\end{quote}

This suggests that intensity and effect vary from person to person and from time to time.

Furthermore, in this essay by Ricoeur, there are justifications for EoCoS form(Appendix F).

\subsection*{ Anticipated effort [appraisal theory]}

\citet{Smith:85} includes "anticipated effort" as one of the dimensions of evaluation.
This is a factor that involves "intensity" and "strength."
For example, sadness has a high score in anticipated effort, suggesting that the type of "strength" is different from "arousal" in \citet{Russell:1980}'s theory.

\section{Word Import(Transplant, Graft) Between Systems(Models)}
\subsection{Hook}
As a general theoretical strategy\citep{Tsuchiya:1986},
we forcus on whether the target element is hooked by the descriptions of model (system), as if being caught on a fishing hook by the model.

\subsection{"Fundamental Concept"}
We set up a criterion to check whether concepts are fundamental. 
The criterion is whether a concept is more fundamental than Hume's three basic relations: resemblance, contiguity, causation.

\subsection{Vicinity of target phenomenon}
If a certain $model_{\,1}$ can hook(pick up) a certain phenomenon, and there is another $model_{\,2}$ 
that can similarly hook that same phenomenon, and if the phenomenon is positioned in the same "Fundamental Concept,"
then we hypothesize that, in the vicinity of that phenomenon, the two models are similar as systems.

\subsection{Unclear concept}
We hypothesize that concepts that cannot conceive of the opposite of a situation within the world become unclear, and we reject to import such concepts(Appendix H).

\subsection{Import(Transplant, Graft) of "intensity"}
The import of the word "intensity," through \citet{Austin:1962}'s use of the word "force," is justified as follows:

\subsubsection*{Step 1}
\citet{Deleuze:1980} provides several examples of Austin's concept of "illocutionary act."
DG uses the word "instantaneousness" many times as a property of "illocutionary act."
Among the examples, we use the example of "judgment."

\subsubsection*{Step 2}
In the "judgment," we examine whether Austin's "illocutionary force"(Appendix B) within "illocutionary act" can be hooked and expressed by EoCoS.

In EoCoS, "force" corresponds to effect of "biased positioning" of "pleasant."
The internal items of EoCoS are bound by "biased positioning" of "pleasant."
"Transformation" in "illocutionary act (=judgment)" is crossing border under "biased positioning" of "pleasant."

\subsubsection*{Step 3}
If step 2 is possible, the DG term equivalent to Austin's "illocutionary force" can be expressed in EoCoS.
In the sections where DG discusses "judgments," the word "force" is not found, but the word "transform" is present.

In the classification of "illocutionary force," Austin discusses "judgments" and states that a judgment "creates a felon."
This corresponds to DG's expression that a judgment "transforms the defendant into a convict."
We avoid saying that Austin's "illocutionary force" is "cause" of DG's "transformation" in a causal chain.
Instead, we regard "illocutionary force" as "production of a certain type of constraint" (Appendix B.4).
We consider "production" to be equivalent to "transformation"(Appendix B.2).
"A certain shape of constraint of preference" $\emph{\textbf{appears in}}$ "transformation."
We provisionally call this "What Appear in Effect."

\subsubsection*{Step 4}
If Austin's "illocutionary force," which is "What Appear in Effect" of "transformation," can be expressed in EoCoS, 
then what is "What Appear in Effect" in DG's model is considered to be expressible in EoCoS.

\subsubsection*{Step 5}
In the works of DG and Deleuze, we examine the relationship between "transformation" and "force."

In "Foucault"\citep{Deleuze:1986} ,  "abstract machine" is described as a "map of relations between forces."
"Abstract machine" is considered to be "immanent cause (Appendix B.5)" of  "assemblage" that realizes the relations contained within "abstract machine."

"What Appear in Effect" of "transformation" in "assemblage" can be considered to be the "map of relations between forces" in the "abstract machine."

If step 4 is possible, then setting "map of relations between forces" as "What Appear in Effect" of "transformation" in our model cannot be considered unreasonable.
This corresponds to an EoCoS deployment.

This paper does not determine whether Deleuze's "force" and Austin's "force" are the same thing.

\subsubsection*{Step 6}
We examine the relationship between "force" and "intensity" in DG and Deleuze. 
In the section\citep{Deleuze:1980} concerning the correspondence between CsO(Corps sans Organes) and Spinoza's "Ethica", 
"intensity" can be interpreted as the degree of "force."
In "Foucault"\citep{Deleuze:1986}, within "abstract machine", a "map of intensity" is juxtaposed  with a "map of relations between forces." 
Here, "intensity" is interpreted as the degree of "force."

\subsubsection*{Step 7}
In step 5, "map of relations between forces" is set in our model.

In step 6, "intensity" is a degree of "force."

Assuming that the word "degree" is a fundamental term with only minor differences in meaning between systems, 
it is not impossible to import "intensity" as the "degree" of "force" in our model.

\section{Other attributes}

\subsection*{Granularity}
The unit of interaction between EoCoSs through montage serves as an indicator of granularity.
However, concepts like "Effect of Structure" are not clearly addressed by natural language.
It is not easy to define the granularity of natural language.

\subsection*{Scope of effect}
In montage of EoCoS, the scope of influence of a demonstrative may be different from what is shorthandly described in natural language.
The scope of influence is different from what is generally considered to be "pointing to."

When considering the threshold of the strength of an influence,
it becomes necessary to precisely define what strength is.
By doing so, we can clearly define the scope of influence at a certain strength threshold.

\subsection*{Not forming oneness(or Unclear individuality)  at the root}
It is conceivable that things that are scattered and less likely to form oneness can produce an effect through an unclear process.
In novels similar to "Fox and Chicken," those that contain several elements like "Don't you know that?" are examples of this type.

\subsection*{Structural stability}

Depending on the types of internal items in EoCoS, the type of montage, and intensity, 
EoCoS and montage allow for the expression of numerous nuances.
However, the range of emotions that appear in daily life are limited within a given culture.
Could this structural stability be explained using terms like "force" and "intensity" ?

\section{Related Works}
We overview theories that put forward the proposition that emotions are constructed.

\subsubsection*{Appraisal theory and Constructivist theory}

Appraisal theories vary in whether they emphasize structure or process , and 
several aspects of the OCC model that focuses solely on structure is compatible with Constructivist approaches \citep{Clore:2013}.
\citet{Scherer:2022} defines Constructivist theories as "presuming that each individual constructs a subjective label to conceptualize a core affect defined by a position in a two-dimensional valence x arousal space," and discusses theory convergence in emotion science.
\citet{Barrett:2007} criticize appraisal dimension, but also points out commonalities\citep{Gross:2011}.

\subsubsection*{Appraisal theory and LLM}
\citet{Tak:2025} analyze where the layer that captures appraisal concepts is located.
\citet{Croissant:2023} add a step to the prompt that contains information designed to allow LLM to appraise the situation to generate emotions of the involved person.
In \citet{Troiano:2023}'s corpus creation, writers first are tasked to remember an event that caused a particular emotion.

\subsubsection*{Emotions development}
Based on the idea that emotions are represented as events themselves, 
\citet{Klinger:2023} argues that dissecting events through emotion role labeling with the help of appraisals is the input for second-level emotion analysis, and that such dissected chains are useful for understanding how emotions develop throughout longer sequences.

\subsubsection*{Individual differences}
\citet{Ong:2021} propose "probabilistic programming" that can explicitly model individual differences and can be composed together in a hierarchical and/or sequential fashion to produce complex phenomena.

\subsubsection*{Intensity(or rating) }
SECEU\citep{Wang:2023}, EQ-Bench\citep{Paech:2023} deal with "intensity."
CoRE\citep{Bhattacharyya:2025}, for example, contains a prompt to ask the amount of "pleasant" in scenarios related to "happiness."

\subsubsection*{Social reality}
In response to the question "Are emotions real?," moving away from the idea that "anger,fear and so on are mere words from folk psychology and should be discarded" ,
 \citet{Barrett:2017a} begun to focus on a different aspect: emotion concepts and emotions have social reality.

\section{Conclusion}
EoCoS is a orientation to avoid contradictions based on the biased positioning of "pleasant."
Within EoCoS, situations that are in a state of contradiction become the starting point for change.
EoS process is avoiding contradictions based on the biased positioning of "pleasant."
When two EoCoSs are in any of three relations(resemblance, contiguity, causation), the action point occurs.
The process is simple. 

As Eisenstein emphasizes, montage involves the "simultaneity" of overlapping elements.
While there is a sequence in the presentation of scenes, the nature of causality is not so much a linear, sequential causality, but rather a structural and simultaneous causality.

\section{Limitations}

We adopt the stance that terms such as EoCoS are just settings that we view as moving elements within the system of our model, and are not asserted as the truth of the world.
However, this does not mean that the predecessors from whom our model borrow concepts did not assert their ideas as the truth of the world.

We reject part of the meaning of the term "difference" which is a core concept in Deleuze's model 
because we reject the import of terms related to ontology,
The problem is, when such a procedure is applied, what is lost and what remains from the word "intensity," or what other problems might arise ?
It remains necessary to examine whether the import of "intensity" is valid.

We try to express "illocutionary force" using EoCoS and montage.
We would like to report on this at a later date.

The intensity of EoCoS varies depending on the type of event.
For demo, a provisional intensity is set.
The intensity should be adjusted to the appropriate level based on subsequent investigations.
The shape of EoCoS is constantly being revised, and the demo shows the current shape.

We would like to classify both EoCoS and natural language into the category of "expression form."
EoCoS is a type of graph from a morphological perspective, and there may be graph-specific constraints.

\section*{Acknowledgments}

Without the Japanese translations of our predecessors' philosophers, our research would not have been possible.
We would like to express our gratitude to the translators and publishers.

\nocite{Smith:85,Clore:2013,Scherer:2022,Bhattacharyya:2025,Wang:2023,Paech:2023,Ong:2021,Tak:2025,Troiano:2023,Klinger:2023,Croissant:2023,Barrett:2017a,Barrett:2007,Gross:2011,Barrett:2017b,Natsume:1908,Deleuze:1968,Deleuze:1994,Deleuze:1980,Deleuze:1987,Deleuze:1986,Deleuze:1988,Deleuze:1983,Foucault:1975,Foucault:1977, NHK:2022, Derrida:1980,Derrida:1987,Strawson:1964,Grice:1967,Grice:1989,Austin:1962,Shibata:2007,Eisenstein:1937,Ricoeur:1950,Ricoeur:1966,Russell:1980,Tsuchiya:1986,Takahashi:2025a,Takahashi:2025b,Takahashi:1992,Derrida:1971,Derrida:1977,Derrida:1972,Derrida:1982,Hume:1739,Hjelmslev:1943,Hjelmslev:1961,Metz:1968,Metz:1991}

\bibliography{custom}

\appendix

\section{Demo of EoS process}
\label{sec:appendix}

We created original material to illustrate EoS process.
In an era when enrollment rates were low, 
scenes in which a teacher visits a student's home to encourage parents can be seen in films 
such as Albert Camus's autobiographical film "Le premier homme." 
While the material we created differs from those scenes, the core of the motif is similar.

The demo may seem to show the process of the characters' mental states,
but we model the mental process of reader(or author) as they read the novel.
The mental state of the characters in a novel is a reflection of the reader's own mental state. 

For the reader, the repetition of a certain word or image is reinforced as they read, 
separate from the inner thoughts of the characters.
However, in demo, reinforcement through similarity is limited to reinforcement between the same characters.
Reinforcement through causality is handled differently from similarity.

These EoS processes are expected to be performed by LLMs.

\section{Illocutionary force}

\citet{Austin:1962} distinguishes between three: "locutionary act", "illocutionary act", and "perlocutionary act".

\begin{quote}
 we  distinguished  the locutionary act  (and 
within it the phonetic,  the phatic,  and the rhetic acts) 
which has a meaning; 

the illocutionary act which has a 
certain force  in  saying  something;  

the  perlocutionary act  which  is  the  achieving  of certain  efects  by  saying 
something. 
\end{quote}

\vskip\baselineskip
\begin{quote}
We  first  distinguished  a  group  of  things  we  do  in 
saying  something,  which  together  we  summed  up  by 
saying we  perform  a  locutionary act,  which  is  roughly 
equivalent  to uttering a certain sentence with  a certain 
sense and  reference,  which  again is roughly  equivalent 
to 'meaning'  in  the traditional  sense.  

Second, we  said 
that we also perform illocutionary acts such as informing, 
ordering, warning, undertaking, i.e. utterances which 
have a certain (conventional) force. 

Thirdly, we may also 
perform  perlocutionary  acts:  what  we  bring  about  or 
achieve by  saying  something,  such as  convincing,  persuading, deterring, and even, say, surprising or misleading.
\end{quote}

\subsection{How it is to be taken}

Austin states the need to "make explicit the precise force of the utterance."

\begin{quote}
For example 'Bull'  or 'Thunder' 
in a primitive language of one-word  utterances could be 
a  warning,  information,  a  prediction.  It  is also  a 
plausible view that explicitly distinguishing the different 
forces  that this utterance might have is a  later achievement of language, and a considerable one;  primitive or 
primary forms of utterance will preserve the 'ambiguity' 
or 'equivocation'  or 'vagueness'  of primitive language in 
this respect; they will not make explicit the precise force 
of the utterance. This may have its uses: but sophistication  and  development  of  social  forms  and  procedures 
will necessitate clarification. But note that this clarification is as much a creative act as a discovery or description! 
It is as much a matter of making clear distinctions as of 
making already existent distinctions clear. 
\end{quote}

The goal of "explicitness" is said to be "how it is to be taken."

\begin{quote}
 Language as such and in its 
primitive stages is not  precise, and it is also not, in our 
sense,  explicit:  precision  in  language  makes it clearer 
what  is  being  said-its  meaning:  explicitness, in  our 
sense, makes clearer the force  of the utterances, or 'how 
(in one sense; see below) it is to be taken'. 
\end{quote}

We can see the relationship between "illocutionary force" and "illocutionary act".

\begin{quote}
 We  said  long  ago  that  we  needed  a  list  of 
'explicit performative verbs';  but in the light of the more 
general theory we now see that what we need is a list of 
illowtionary forces  of an utterance. The old  distinction, 
however, between primary and explicit performatives will 
survive the sea-change  from the performative/constative 
distinction  to the  theory  of  speech-acts  quite  successfully. 
For we have since seen reason to suppose that the 
sorts of test suggested for the explicit performative verbs 
(to  say ... is to ...) will do, and in fact do better 
for  sorting out those verbs which make  explicit,  as we 
shall now say, the illocutionary force of an utterance, or 
what  illocutionary  act  it  is  that  we  are  performing  in 
issuing that utterance. 
\end{quote}

\subsection{"Transform and produce", "transform versus referent"}
The description by \citet{Derrida:1977} in 1971 is consistent with \citet{Deleuze:1980}.
And this description suggests that the word "transform" is a basic concept.
and suggests that "transform" and "produce" are words of the same level.

\begin{quote}

3)  As  opposed  to  the  classical  assertion,  to  the  constative  utterance,  the 
performative does not have its referent (but here that word is certainly no longer 
appropriate, and this precisely is the interest of the discovery) outside of itself or, 
in any event, before and in front  of itself.  It does  not describe something that 
exists outside of language and prior to it.  It produces or transforms a situation, it 
effects; and even if it can be said that a constative utterance also effectuates something and always transforms a situation, 
it cannot be maintained that that constitutes its internal structure, its manifest function or destination, as in the case of 
the performative.

\end{quote}

When comparing the word "transform" in Austin's, Derrida's, and Deleuze's systems, how can we say that they "have these kinds of differences"?

\subsection{Production, effect and force}

In 1971, \citet{Derrida:1977} juxtaposes the communicating "force" with the production of "effect."

\begin{quote}

2) This  category of communication is  relatively new. Austin's notions  of illocution  and perlocution  do  not  designate  the  transference  or  passage  of a 
thought-content, but, in some way, the communication of an original movement 
(to be defi ned within a general theory  of action), an operation and the production of an effect. Communicating, in the case of the performative, if such a thing, 
in all rigor and in all purity, should exist (for the moment, I am working within 
that hypothesis and at that stage of the analysis), would be tantamount to communicating a force through the impetus [impulsion] of a mark. 

\end{quote}

\subsection{Effect [Austin]}

Regarding the word "effect," it seems necessary to distinguish between four.
Three "effects" in "illocutionary act" 
and "effect" in "perlocutionary act."

"Securing uptake," "taking effect," and "inviting responses" are the three of "illocutionary act."

\begin{quote}
Thus we  distinguished  the locutionary act  (and 
within it the phonetic,  the phatic,  and the rhetic acts) 
which has a meaning; the illocutionary act which has a 
certain force  in  saying  something;  the  perlocutionary 
act  which  is  the  achieving  of certain  efects  by  saying 
something. 
We  distinguished  in  the  last  lecture some senses of 
consequences and effects in these connexions, especially 
three  senses  in  which  effects  can  come  in  even  with 
illocutionary acts, namely, securing uptake, taking effect, 
and inviting responses. In the case of the perlocutionary 
act we  made  a rough  distinction between achieving an 
object  and  producing  a  sequel.  Illocutionary  acts  are 
conventional acts:  perlocutionary  acts  are  not  conventional.
\end{quote}

A more detailed description is below.

\begin{quote}
I have so far argued, then, that we can have hopes of 
isolating the illocutionary act from the perlocu tionary as 
producing consequences, and that it is not itself a 'consequence'  of the locutionary act. Now, however, I must 
point  out that the illocutionary act as distinct from the 
perlocutionaryis connected with the production of effects 
in certain senses : 

(I) Unless a certain effect is achieved, the illocutionary 
act will not have been happily,  successfully performed. 
This is to be distinguished from saying that the illocutionary act is the achieving of a certain effect.  I cannot 
be said to have warned an audience unless it hears what 
I say and takes what  I say in a certain sense. An effect 
must be achieved on the audience if the illocutionary act 
is  to be  carried out.  How  should  we  best  put it here? 
And how can we limit it ?  Generally the effect amounts 
to bringing about the understanding of the meaning and 
of  the force of  the locution.  So the performance of  an 
illocutionary act involves the securing of uptake. 

(2)  The illocutionary act 'takes effect'  in  certain ways, 
as  distinguished  from  producing  consequences  in  the 
sense of bringing about states of  affairs in the 'normal' 
way, i.e.  changes in the natural  course of  events.  Thus 
'I name this ship the Qaeen Elizabeth'  has the  effect of 
naming or christening the ship ; then certain subsequent 
acts  such as referring  to it as the  Generalissimo Stalin 
will be out of order. 

(3)  We have said that many illocutionary acts invite by 
convention a  response  or  sequel,...

\end{quote}

\vskip\baselineskip
\begin{quote}
So here are three ways in which illocutionary acts are 
bound  up  with  effects;  and these are all  distinct  from 
the  producing  of  effects which  is  characteristic  of  the 
perlocutionary act. 
\end{quote}

Grice's reference to "uptake" is given in Appendix B.8.

We now focus on the "taking effect" using Austin's example of ship naming.
If we were to express the temporal aspect of the "transformation" "in" "illocutionary act" in \citet{Deleuze:1980}'s words, 
it would be "instantaneousness."
If we write "by" instead of "in," 
we will have to make another disclaimer in relation to the "perlocutionary act," 
so it seems that writing "in" will avoid any problems.

\vskip\baselineskip

\vskip\baselineskip
We Translate the example of ship naming into EoCoS:

Before the naming, there was "biased positioning" of "pleasant" created by "contiguity" 
between the people involved, the citizen, the ship, the specific name, and "pleasant."
The internal items of EoCoS are bound by "biased positioning" of "pleasant."
In EoCoS, "force" corresponds to "biased positioning" of "pleasant."

After the naming, "biased positioning" of "pleasant" has a new positioning with the internal items in EoCoS.
The act of naming results in an EoCoS with a "biased positioning" in which it would be "unpleasant" to call or refer to the object that has been named by any other name.
The term "calling" occupies one position as an internal item of EoCoS.
After the naming, new "forces" appears.

In EoCoS, "transformation" is a new appearance of "biased positioning" of "pleasant" or crossing border under "biased positioning" of "pleasant."
In EoCoS, "force" is not something that is located at the beginning of a causal chain.
EoCoS as a whole expresses "force."

\vskip\baselineskip
Austin explains the "classes of utterance, classified according to their illocutionary force."
Austin lists specific examples of verbs in each class.
We continue our discussion, taking the verbs into consideration.

\begin{quote}
(I)  Verdictives. 

(2)  Exercitives. 

(3)  Commissives. 

(4)  Behabitives (a shocker this). 

(5)  Expositives. 

\end{quote}

\vskip\baselineskip
Here, we focus on Austin's interpretation of Derrida(Appendix B.2,3) and 
note Derrida's juxtaposition of the following words: 
transform, produce, production, effect, and force.

Our current question is how to state "illocutionary force" without using EoCoS.
We try to replace "biased positioning" of "pleasant" in EoCoS with the word "preference."

We interpret the accompanying "taking effect" in "illocutionary act" as equivalent to "illocutionary force," 
and consider "illocutionary force" to be:

\begin{itemize}
 \item a certain shape derived from the way in which preference is applied
 \item producing a certain shape under constraint of preference
 \item production of a certain type of constraint
 \item a certain form that is constrained by preference
 \item a certain form shaped by constraint of preference
\end{itemize}

\vskip\baselineskip
"A certain shape of constraint of preference" appears in "transformation."
We provisionally call this "What Appear in Effect."

\vskip\baselineskip
We think that "illocutionary force" is not a "cause" of "transformation" in a causal chain.
\citet{Deleuze:1986} uses the word "immanent cause" to describe a situation equivalent to this situation.

\subsection{Immanent cause}
\citet{Deleuze:1986} uses the word "immanent cause" to describe the relationship between "diagram (=abstract machine)" and "assemblages."

\begin{quote}
None the less, the diagram acts as
a  non-unifying  immanent  cause  that  is coextensive  with  the
whole social field: the abstract machine is like the cause of the
concrete  assemblages  that  execute  its  relations;  and  these
relations  between forces  take place  'not above' but  within  the
very  tissue of the  assemblages  they  produce.
What  do we mean  here  by immanent  cause?  It  is a  cause
which is realized, integrated and distinguished in its effect.  Or
rather  the immanent  cause  is realized,  integrated  and  distinguished  by  its  effect.  In  this  way  there  is  a  correlation  or
mutual  presupposition  between  cause  and  effect,  between
abstract  machine  and  concrete  assemblages  (it  is  for  the
latter  that  Foucault  most  often  reserves  the  term
'mechanisms'). 
\end{quote}

\subsection{Inviting responses}
In naming ships, what we describe below falls into category A or B in Austin.

A: The third effect "inviting responses" in "illocutionary act" 

B: Effect in "perlocutionary act"

\vskip\baselineskip
Without an official name for the ship, it is impossible to refer to this luxurious and distinctive object by its proper name.
The people involved are in a state of contradiction.
Once the official naming is complete, questions will arise, for example, as to whether the voyage plan is worthy of the name.
The ship is affected by at least several forces: the "force" that comes from the ship's size, equipment, and luxury; 
the "force" of the country; the "force" that comes from the name; the "force" of the society of ship crews; and the "force" of the society of voyage planners.
With the ship, these forces overlap.
If a public figure in public were to mistakenly call the ship by another name,
it would have a very different effect than calling a flower by the wrong name.

\vskip\baselineskip
EoCoS is created for each different type of subject. 
If the subject type is, for example, a citizen, and the items inside EoCoS are different,a new EoCoS is set up.
Also, "intensity" of EoCoS may differ.
This type of setup seems directly related to what \citet{Deleuze:1986} calls "microsociology."

\vskip\baselineskip
However, Grice's "effect" is different from Austin's.
This issue is related to the question of how to think about the scope of "conventional act."

\subsection{Force versus truth}

\citet{Derrida:1971,Derrida:1977} feels a kinship with Austin's attitude towards "truth."

\begin{quote}

4) Austin was obliged to free the analysis of the performative from the author-
ity of the truth value, from the true/false opposition, at least in its classical form, 
and to substitute for it at times the value of force, of difference of force (illocutionary  or perlocutionary force). 
(In this line of thought, which is nothing less than 
Nietzschean, this in particular strikes me as moving in the direction of Nietzsche 
himself, who oft en acknowledged a certain affinity for a vein of English thought.) 

\end{quote}

\subsection{Effect[Grice]}
Grice who studied closely\citep{Takahashi:2025a} with Austin uses the word "effect" as a central concept in semantics.
Grice's use of the word "effect"  is different from Austin's.

In "Prolegomena(1967)," \citet{Grice:1989} mentions "uptake", the first of Austin's three "effects" that accompany "illocutionary act."

\begin{quote}
The most general complaint, which comes from Strawson, Searle, 
and Mrs. Jack, seems to be that I have, wholly or partially, misidentified the intended (or M-intended) effect in communication; 
according to me it is some form of acceptance (for example, belief or desire), 
whereas it should be held to be understanding, comprehension, or (to use an Austinian designation) "uptake."

One form of the cavil (the 
more extreme form) would maintain that the immediate intended target is always "uptake," though this or that form of acceptance may 
be an ulterior target; a less extreme form might hold that the immediate target is sometimes, but not invariably, "uptake." 

\end{quote}

\section{Assemblage and illocutionary act}

\citet{Deleuze:1980} juxtapose "assemblages of enunciation" with "illocutionary act."

\begin{quote}
The order-words or assemblages of enunciation in a given society 
(in short, the illocutionary) designate this instantaneous relation between 
statements and the incorporeal transformations or noncorporeal attributes 
they express. 
\end{quote}

\section{Abstract machine and assemblage}

\citet{Deleuze:1980} describe the relationship between "abstract machine" and "assemblage."

\begin{quote}
For a true abstract machine pertains to an assemblage in its entirety: it is defined as the diagram of that assemblage.
\end{quote}

\section{Map of relations between forces}

\citet{Deleuze:1986} describe the relationship between "abstract machine" and "map  of  relations  between  forces."

\begin{quote}
What is a diagram ? It  is a display of the relations  between
forces  which constitute power  in  the  above  conditions:
The  panoptic  mechanism  is not  simply a  hinge,  a  point of
exchange between a mechanism of power and  a function;  it
is a way  of making power  relations functions  in a  function,
and  of making a  function  through these power  relations. 
We  have  seen  that  the  relations  between  forces,  or  power
relations,  were  microphysical,  strategic,  multipunctual  and
diffuse,  that  they  determined  particular  features  and  constituted  pure  functions.  
The  diagram  or  abstract  machine  is the  map  of  relations  between  forces,  a  map  of  destiny,  or
intensity, which proceeds by primary non-localizable relations
and at every moment passes through every point, 'or rather in
every relation from  one point to another'.

\end{quote}

\section{Anticipation of pleasure}

"Ideal image" in our model has an affinity with \citet{Ricoeur:1950}'s concept of "anticipation of pleasure and pain."
EoCoS can be seen as a representation of "experience of need as lack."

\begin{quote}
Pleasure  in  fact  enters  motivation  through  the  imagination:  thus  it  is  a  moment  in  desire. 
Desire  is  the  present  experience  of  need  as  lack  and  as  urge, 
extended  by  the  representation  of  the  absent  object  and  by  anticipation  of  pleasure.  
\end{quote}

The words "common form" and "close similarity" below are parallel to 
our description "If ideal image and actual image match, the result is pleasant, and if they don't match, the result is unpleasant."

\begin{quote}
on the  human  level,  organic  life  is  undoubtedly  a  cluster  of  heterogeneous  demands,  revealing  discordant  values. 
This  ambiguity  of  organic  life  is  what  is  really  at  stake  in 
this  analysis  which  would  otherwise  often  appear  as  a  rather uncertain  unraveling. 
As a counterpart  to this, we shall show  that 
it  is  always  anticipating  imagination  which  transmutes  the  multiple sources  of motives issuing from  
the body and which  bestows a common form,  a form  lending itself  to conventional  value judgments:  
"this  is  good,  that  is  bad." 

\end{quote}

\vskip\baselineskip
\begin{quote}
 imagination  profoundly  transforms  this  situation,  instituting a  close similarity  
between  anticipated  pain  and  anticipated pleasure.  
Imagined  pleasure  is  called  desire - imagined  pain  is called fear.  
But while desire extends  need which itself  anticipates pleasure,  fear  reverses  the  order  of  precedence  between  action 
and  the  painful  encounter.  
Fear  can  precede  and  ward  off  the threat  just  as  need  and  desire  preceded  and  sought  pleasure.  
In this  way  imagination  likens  fear  to  a  negative  desire  and  fear 
reveals  pain  as  evil,  that  is,  as  the  opposite  of  the  good. 

\end{quote}

\section{Conventional act}
\subsection{Strawson}

\citet{Strawson:1964} urges careful distinctions to be made around "linguistic conventions," and Grice states the same thing, as shown below.

\begin{quote}
 First, we may agree (or not dispute) that any
 speech act is, as such, at least in part a conventional act. The
 performance of any speech act involves at least the observance or
 exploitation of some linguistic conventions, and every illocutionary
 act is a speech act. But it is absolutely clear that this is not the
 point that Austin is making in declaring the illocutionary act to
 be a conventional act. 
\end{quote}

Below is a passage where Strawson uses the phrase "force of an objection."
Strawson uses the word "force" in agreement with Austin's concept, but distances himself from Austin's use of the word "conventional."

\begin{quote}
 In the course of a philosophical discussion
 (or, for that matter, a debate on policy) one speaker raises an
 objection to what the previous speaker has just said. X says (or
 proposes) that p and r objects that q. r's utterance has the force
 of an objection to X's assertion (or proposal) that p. But where
 is the convention that constitutes it an objection ? That r's utterance
 has the force of an objection may lie partly in the character of the
 dispute and of X's contention (or proposal) and it certainly lies
 partly, in r's view of these things, in the bearing which he takes
 the proposition that q to have on the doctrine (or proposal) that
 p. But although there may be, there does not have to be, any
 convention involved other than those linguistic conventions
 which help to fix the meanings of the utterances.
\end{quote}

A one-off political example that \citet{Deleuze:1980} give as illocutionary acts may support Strawson.
This is an example of an utterance that creates a new, unfamiliar political object that did not exist before.

\subsection{Grice}

In "Prolegomena(1967)," \citet{Grice:1989} also urges careful distinctions to be made around "linguistic conventions."

\begin{quote}
My impression is that Searle (like Austin) thinks of 
speech-acts of the illocutionary sort as conventional acts, the nature 
of which is to be explained by a specification of the constitutive rules 
which govern each such act, and on which the possibility of performing the act at all depends. An infraction of one of these rules may 
mean (but need not mean) that an utterance fails to qualify as a specimen of the appropriate type of speech-act; it will at least mean that 
the utterance is deviant or infelicitous. 

Now, while some speech-acts (like promising, swearing, accepting 
in marriage) may be conventional acts in some such sense as the one 
just outlined, and while remarking is no doubt a conventional act in 
some sense (since it involves the use of linguistic devices, which are in 
some sense conventional), I doubt whether so unpretentious an act as 
remarking is a conventional act in the above fairly strong sense. This 
issue cannot be settled in advance of an examination of the character 
of speech-acts and of the meaning of the phrase "conventional act." 

\end{quote}

\section{Unclear concept}
\citet{Deleuze:1968}'s model includes an ontology, centered on the core term "difference," that is close to Bergson's concept of "duration."
Furthermore, \citet{Deleuze:1980}'s model also includes Spinoza's concept of "the univocity of being."

We hypothesize that the difficulty in understanding these ontologies stems less from the profundity of philosophy itself and more from the inherent ambiguity of the concepts themselves.
\citet{Takahashi:1992} develop Saussure's hypothesis that the origin of meaning lies in "perception of difference" and proposes the following hypothesis:
When a concept has no counterpart to which it can perceive a difference, or when one cannot conceive of the opposite of a situation within the world, the concept become unclear.

For example, the event of a book being on a table allows for the opposite statement.
We can say things like, "The book is not on the table. It's in the car," 
and have no trouble imagining and understanding the opposite situation.

We hypothesize that concepts that cannot conceive of the opposite of a situation within the world become unclear, and we reject to import such concepts.

The problem is, when such a procedure is applied, what is lost and what remains from the word "intensity," or what other problems might arise?

\section{Montage}
The concept of "montage," which is said to have originated from Kuleshov's experiments, is surrounded by much debate in film theory.

\citet{Metz:1968} states: 

\begin{quote}
Eisenstein soon let his own mind be conquered by the desire to conquer other minds, 
and he became the leading theoretician of the "montage or bust" approach."
and that "he loudly asserted that it was montage simply because Dickens, Leonardo da Vinci, and about twenty other people combined two themes, two ideas, or two colors.
\end{quote}

\vskip\baselineskip
\begin{quote}
It was enough  that  Dickens,  Leonardo  da  Vinci,  or  any  number  of  others 
combined two themes,two ideas, or two colors for Eisenstein to discover montage; 
the most obvious pictorial juxtaposition, the most 
properly literary effect of composition,were, to hear him, prophetically precinematographic.
All is montage. There is something relentless, almost embarrassing at times, in Eisenstein's refusal to admit 
even the smallest place for continuous flows of creation; all he can 
see anywhere are pre-fragmented pieces, which ingenious manipulation  
will  then  join  together.  Furthermore,  the  manner  in  which  he 
describes the creative work of all those he enlists as his forerunners 
does not fail, in certain truly improbable passages, to contradict even 
the slightest likelihood of any psychogenesis of creation.
\end{quote}

\citet{Shibata:2007} focuses on Bazin's critique of montage and states that "the purpose of the technique of montage is to produce a certain effect through a chain of cuts," 
and then notes that Bazin argues that "montage is not the essence of film, but merely one technique," and considers realism to be at the core of Bazin's film aesthetics.

Building on Bergson's ideas, \citet{Deleuze:1983} develops a unique perspective, stating that "montage is the combination of three types of movement images."

Our concept of montage currently focuses on the ups and downs of "strengthen" and "intensity." 
We anticipate examining the import of words and concepts in the context of Eisenstein's "interaction between images." 
When considering the import of Eisenstein's concept of montage, 
a general theoretical framework, word import between systems, seems to emerge.

\section{Pseudo-expression of will}

In order to possess a pseudo-expression of will, 
in addition to "orientation to avoid unpleasant situations," the following also seems necessary.

That is, there's no escaping the "unpleasantness" that's here.
Humans cannot easily escape from any unpleasant that arise for themselves or others as a result of their own actions.
These unpleasant are attributed to the self that exists here and now.
Machines are fundamentally incapable of possessing the subjective feeling of "unpleasantness."
And even if we were to assume that they could somehow simulate this unpleasantness, there is another problem.
Machines cannot possess the same kind of inseparability between the self and unpleasantness that exists in humans.
In fact, even before discussing machines, we cannot fully understand or articulate what "the inseparability between the self and unpleasantness in humans" actually means.

\section{Intensity [Bergson]}
\citet{Deleuze:1968} refutes Bergson's criticism of "intensity."
Bergson argued that differences in emotional intensity are differences in nuance, and opposed expressing the intensity of emotion in terms of "quantity."
Deleuze countered Bergson through what he called transcendental investigation\citep{Takahashi:2025b}.

\section{Effect [Derrida]}

 The concept of "effect of structure" is used in French structuralism.
Derrida, in particular, frequently uses the word "effect."
We owe our attention to the word "effect" to Derrida.
Derrida redefined the term "structure."
Derrida uses the phrase "effect of diff\'{e}rance." 
"Diff\'{e}rance" is considered to be closely related to Heidegger's "ontological difference." 
Therefore, we avoid using the term "diff\'{e}rance."
The reasons for avoiding ontology words are explained in Appendix H(Unclear Concept).
Aware of Derrida's criticism of "presence," we use the term "effect of structure." 
The difference between "diff\'{e}rance" and "structure" is concisely expressed in 
the phrase "the gram as a new structure of nonpresence (le gramme comme nouvelle structure de la non-presence)" which Kristeva stated during a conversation with \citet{Derrida:1972}.
The word "gram" is close to Derrida's word "\'{e}criture."

\end{document}